\documentclass{article}

\usepackage[preprint]{nips_2018}
\usepackage{natbib}

\usepackage{caption}
\usepackage{subcaption}
\usepackage{amsmath}
\usepackage{mathtools}
\usepackage{amsthm}
\usepackage{physics}
\usepackage{caption}
\usepackage{subcaption}
\usepackage{multirow}
\usepackage{makecell}
\usepackage{url}            
\usepackage{booktabs}       
\usepackage{amsfonts}       
\usepackage{nicefrac}       
\usepackage{microtype}      
\usepackage{algorithm}
\usepackage{algorithmic}
\usepackage{graphicx}
\usepackage{makecell}
\usepackage{stmaryrd}
\usepackage{hyperref}       

\DeclareMathOperator{\argmax}{\arg\max}
\DeclareMathOperator{\argmin}{\arg\min}
\newcommand{\method}{{AbdGen}}
\newcommand{\conabd}{ConMetaAbd}
\newcommand{\vecabd}{AttVerify}

\newcommand{\citepsec}[1]{Sec.~\ref{#1}}

\newcommand{\citepeq}[1]{Eq.~\ref{#1}}

\newcommand{\citepfig}[1]{Fig.~\ref{#1}}
\newcommand{\citeptab}[1]{Tab.~\ref{#1}}

\title{Generating by Understanding: Neural Visual Generation with Logical Symbol Groundings}

{
  \author{Yifei Peng$^1$ \quad Zijie Zha$^1$ \quad Yu Jin$^1$ \quad Zhexu Luo$^3$\thanks{Work done as a visiting student in State Key Laboratory of CAD\&CG, Zhejiang University.} \quad Wang-Zhou Dai$^2$ \quad \textbf{Zhong Ren}$^1$ \\ \textbf{Yao-Xiang Ding}$^1$\thanks{Corresponding author.} \quad \textbf{Kun Zhou}$^1$ \\[1.25ex]
$^1$State Key Laboratory of CAD\&CG, Zhejiang University \\
$^2$National Key Laboratory for Novel Software Technology, Nanjing University \\
$^3$Department of Computer and Information Science, University of Pennsylvania \\[1.25ex]
\texttt{\{pengyf,zjzha,jinyu99\}@zju.edu.cn,zhexuluo@gmail.com,daiwz@lamda.nju.edu.cn}\\
\texttt{
renzhong@zju.edu.cn,dingyx.gm@gmail.com,kunzhou@acm.org}
}

\begin{document}

\maketitle

\begin{abstract}
Making neural visual generative models controllable by logical reasoning systems is promising for improving faithfulness, transparency, and generalizability. We propose the Abductive visual Generation (AbdGen) approach to build such logic-integrated models. A vector-quantized symbol grounding mechanism and the corresponding disentanglement training method are introduced to enhance the controllability of logical symbols over generation. Furthermore, we propose two logical abduction methods to make our approach require few labeled training data and support the induction of latent logical generative rules from data. We experimentally show that our approach can be utilized to integrate various neural generative models with logical reasoning systems, by both learning from scratch or utilizing pre-trained models directly. The code is released at \url{https://github.com/future-item/AbdGen}.   

\end{abstract}

\section{Introduction}
\label{sec:intro}
Following the great attention in research and application on neural visual generative models~\citep{bond2021deep,kingma2013auto,goodfellow2014generative,ho2020denoising}, improving their controllability under external control signals has become a fundamental research problem. To address this challenge, conditional generation techniques have been developed, enabling strong controllability based on various contextual information~\citep{mirza2014conditional,karras2019style,zhang2023adding}. As an alternative, in this work, we focus on building a logic-integrated neural visual generative model, which is the integration of a neural generative model and a logical reasoning system. In this model, the neural generation process is controlled by the grounding values of symbolic concepts in the logical reasoning system. Such a model is particularly suitable for the situation where the high-level semantics of the generation results need to follow rigorous logical regularities, which are hard to encode by neural models but can be easily modeled as the logical background knowledge in the reasoning system. Compared with purely neural models, the logic-integrated model has the following advantages: 1) faithfulness, such that the semantics of generation results are strictly legal under logical verification; 2) transparency, such that the logical reasoning process is conducted in the white-box reasoning system, which is checkable by human users; 3) generalizability, such that by changing the logical background knowledge in the reasoning system, a learned model is able to generate instances following significantly different semantics from those in the training data, which is fundamentally difficult to realize by purely neural models. 

The critical technical aspect in building such a model is symbol grounding learning, which is training a neural symbol grounding module, a part of the neural generative model, to represent logical symbolic concepts. After training, the module can generate neural embeddings with symbolic meanings determined by logical reasoning. The embeddings are then utilized by the neural model to obtain the final visual outputs. We propose a vector-quantized symbol grounding module, together with an adversarial training method for conducting its optimization. Compared with the pioneer work VAEL~\citep{misino2022vael}, the only previous research on logic-integrated visual generation to our knowledge, our approach achieves the more robust disentanglement between symbolic and subsymbolic neural embeddings, hence improving the controllability of logical symbols over generation. 

Furthermore, following previous research~\citep{misino2022vael}, we consider the situation in which the training data follow from a set of logical generative rules. For learning the logic-integrated generative model in this scenario, we introduce two novel symbol grounding learning strategies based on the abductive learning framework~\citep{zhou2019abductive,dai2019bridging}. The target is to address two technical challenges that no previous research tackles with: 1) learning from few labeled data, which is essential when acquiring instance-level symbol grounding annotations leads to high cost; 2) supporting the induction of latent logical generative rules from data, which can be utilized to guide the testing-stage generation process. In the scenario of learning without rule induction, an attention-based symbol verification mechanism is proposed to significantly boost the efficiency of symbol grounding learning. In the scenario of learning with rule induction, we propose the method of contrastive meta-abduction to conduct precise logical rule induction in companion with the process of symbol grounding learning. Experimental results verify that our Abductive visual Generation (AbdGen) approach can be integrated with various neural generative models by learning both from scratch or utilizing pre-trained generation models directly, serving as a systematic solution for building logic-integrated neural visual generative models. 

\section{Related Work}
\label{sec:rw}
\subsection{Neurosymbolic Visual Generation}

Despite some existing work~\citep{jiang2021generative,feinman2020generating,gothoskar20213dp3}, there are not many previous attempts at knowledge-based neural visual generation, especially the utilization of logical reasoning systems. The only exception is VAEL~\citep{misino2022vael}, which pioneers the study of logic-integrated neural visual generation. VAEL integrates the DeepProblog framework~\citep{manhaeve2018deepproblog} with variational autoencoders, achieving impressive ability to generate visual objects based on logical rules, as well as strong generalization when training and testing logical rules differ. On the other hand, VAEL still assumes that sufficient instance-level labels are provided for symbol grounding learning. Meanwhile, no mechanism is provided for performing rule induction. In our view, learning from few labeled data and conducting rule induction both require highly efficient mechanisms of grounding searching and logic program synthesis, which is not the advantage of DeepProblog. The basic mechanism of DeepProblog is to conduct logical reasoning by calculating possible world probabilities, which may involve an intractable calculation of joint distributions. In comparison, abductive learning has the advantage of doing more direct searching over the logic program space, making effective symbol grounding and rule learning from limited labeling information realizable.

\subsection{Symbol Grounding Learning}

The grounding of neural representations to causal variables has become a key research topic recently~\citep{scholkopf2021toward}, and symbolic grounding learning has been studied for SATNet~\citep{topan2021techniques} in addressing satisfiability problems. There are also some pioneering studies on symbol grounding learning for complex visual objects, such as~\citet{huang2022multi,hong20223d,hsu2023ns3d,hsu2024s}. However, most existing research focuses on significantly different tasks from ours, such as visual question answering. The related study is rarely for neural visual generators integrated with logical reasoning systems, except for VAEL. 

On the other hand, our work is built on the abductive learning (ABL) framework~\citep{zhou2019abductive,dai2019bridging}, which is a suitable choice for symbol grounding learning. ABL is a neurosymbolic learning framework that unifies the subsymbolic learning model and symbolic reasoning by employing logical abduction. In the pioneering work of~\citet{dai2019bridging}, ABL is proposed to address the weakly supervised classification problem where the training data are not labeled, and a logical reasoning system is available to generate abduced labels for learning the neural classification model. 
Subsequently, in~\citet{dai2020abductive}, MetaAbd is proposed to further enhance ABL with the ability to learn new logic programs during the abduction process based on meta-interpretative learning~\citep{muggleton2017meta}. Both these approaches propose promising ways to address symbol grounding of neural visual generators, but there exist significant technical challenges: 1) Latent presentations in neural visual generators are usually multidimensional representation vectors instead of simple classification labels, which involve more complicated mechanisms for training; 2) In visual generation tasks, the logical reasoning systems need to model the generative process rules instead of classification criterion, which is usually more complicated, making the time cost of abduction a significant issue. A new abduction strategy must be designed that can utilize the prior information from neural models to improve abduction efficiency; 3) The logical rule learned for generation should be as precise as possible, since generation usually requires more information than classification. As a result, a stronger rule induction method should be proposed.

\section{Logic-Integrated Neural Visual Generation}
The target of our work is to build a logic-integrated visual generative model, which integrates a neural visual generator and a logical reasoning system. The overview of the model is illustrated in \citepfig{fig:model}. 

\begin{figure*}[t]
\centering
\includegraphics[width=.9\linewidth]{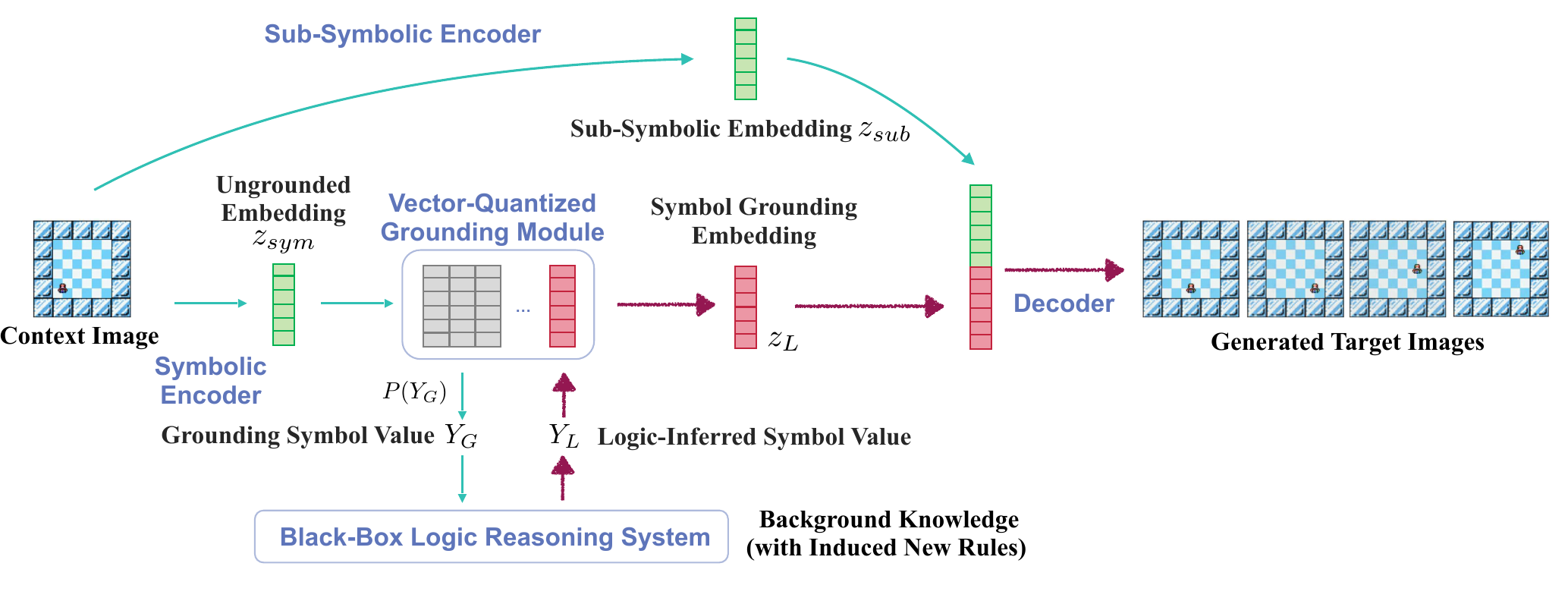}
\caption{Illustration of the logic-integrated neural visual generation model.}
\label{fig:model}
\end{figure*}

\subsection{Model Design} 
The neural visual generator $(E_{sym}, E_{sub}, V, G)$ consists of a symbolic encoder $E_{sym}$, a sub-symbolic encoder $E_{sub}$, a vector-quantized grounding module $V$, and a decoder $G$. The encoders $E_{sym}$ and $E_{sub}$ are responsible for transforming context image $x$ into ungrounded latent embedding $z_{sym}$, which preserves symbolic information but is ungrounded, and sub-symbolic embedding $z_{sub}$, which preserves symbol-independent visual features. 
The vector-quantized grounding module $V$ is responsible for mapping $z_{sym}$ into grounded embeddings $z_{G}$, and getting their corresponding grounding symbol values $Y_G$. 
This encoding process can be expressed with
\begin{equation}
z_{sym} \leftarrow E_{sym}(x),\; (z_{G}, Y_G) \leftarrow V(z_{sym}),\; z_{sub} \leftarrow E_{sub}(x).
\label{eq:gen}
\end{equation}
The grounding module $V$ can also take logic-inferred symbol values $Y_L$ as input and retrieve their corresponding grounded embeddings $z_L$. Given $z_{L}$ and $z_{sub}$ as inputs, the decoder $G$ generates the target images $\hat x \leftarrow G(z_{L}, z_{sub})$.

{\noindent\bf Logical reasoning system}. The logical reasoning system has its internal {\it background knowledge set} $B$, which are logical clauses that define a set of logical facts, as the basis of reasoning. The rules in $B$ form the strict constraints for generation, which are task-related and are mostly user-defined. Furthermore, in the scenario of learning with rule induction, {\it new logical rules can be induced from the data to augment $B$}. 
We assume that the logical reasoning system is not necessary to be differentiable and can not be directly integrated with the neural part to conduct end-to-end learning. As a result, our approach can be implemented with various logical reasoning systems. 

{\noindent\bf Vector-quantized grounding module}. The grounding module $V$ is a vector-quantized model~\citep{van2017neural}. $V$ consists of a set of dictionaries $\mathcal Z$ for all symbols. Each dictionary consists of a set of learnable embeddings $z_Y\in\mathcal Z$, each representing a value $Y\in\mathcal Y$ for the symbol, where $\mathcal Y$ is the value space. $V$ majorly has two functionalities: 1) retrieving the grounded embedding $z_G$ and its grounding symbol value $Y_G$ based on the ungrounded embedding $z_{sym}$; 2) retrieving a specific grounding embedding $z_L$ based on the symbol value $Y_L$. The first operation is done by calculating the probabilities of the symbol values based on embedding distances:
\begin{equation}
 P(Y) = \exp(-\|z_{sym}-z_Y\|)/D,
 \label{eq:metric}
\end{equation}
in which $D=\sum_{Y\in\mathcal Y}\exp(-\|z_{sym}-z_Y\|)$. Then $Z_G$ and $Y_G$ are decided as the ones with the highest probability. The second operation is trivial: $z_L$ is fetched directly from the dictionary according to $Y_L$.

\subsection{Generation Process}
As illustrated in \citepfig{fig:model}, the model takes a context image $x$ as input. The model then conducts the neural encoding process to extract the {\it grounding symbol values} $Y_G$.  
Afterwards, $Y_G$ is sent to the logical reasoning system, which is equipped with a set of {\it logical generative rules} as its background knowledge. Based on logical reasoning, a set of {\it logic-inferred symbol values} $Y_L$ is generated. Finally, the model performs the neural decoding process to obtain the target images $\hat x$, which should be consistent with the symbol-independent visual style of the context image, as well as the symbol values decided by the logical generative rules. For example, in \citepfig{fig:model}, given the context image and the symbolic rule {\it Mario moves with right priority followed by up priority}, a sequence of target images are generated. The symbol values are the positions of Mario on the map.

\section{Visual Generative Abductive Learning}
\label{sec:method}
\begin{figure*}[t]
\centering
\includegraphics[width=.9\linewidth]{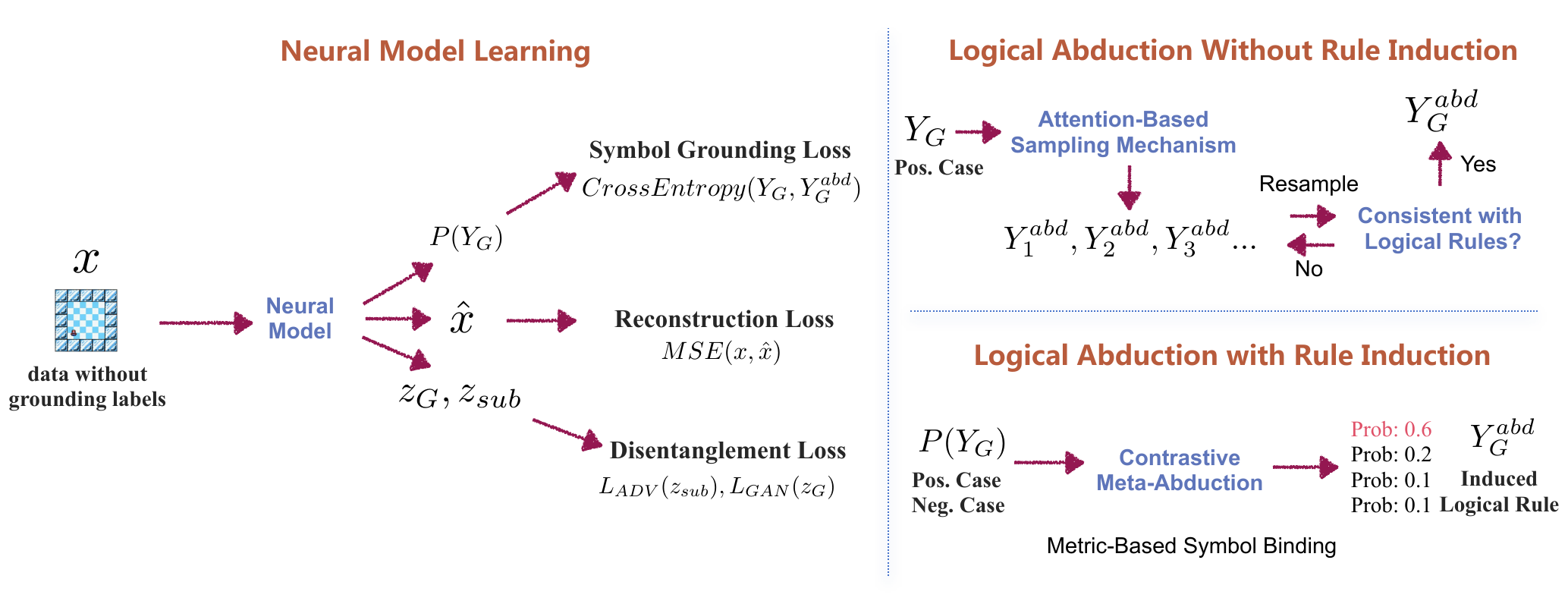}
\caption{The visual generative abductive learning process.} 
\label{fig:learn}
\end{figure*}
In this section, we introduce the learning approach for the logic-integrated neural visual generation model. The learning process has three perspectives: 1) training of the neural encoding part ($E_{sym}, V$), which should accurately extract symbol grounding values from the context image; 2) training of the neural decoding part ($E_{sub}, G$), which should generate high-quality images following logic-inferred symbol values and symbol-independent visual style of the context image; 3) training of the logical reasoning part, which should be able to induce latent logical generation rules from data, such as the {\it right priority rule} illustrated in \citepfig{fig:model}, to augment the reasoning system. We consider two learning scenarios, {\it learning without rule induction} and {\it learning with rule induction}. For the first situation, the background knowledge set $B$ in the logical reasoning system is complete. For the second, $B$ is incomplete, so it is necessary to induce the missing rules from the data. We also introduce the following assumptions about the training data set $X$.
\begin{itemize}
    \item 
    The dataset is formed by a set of {\it cases}. A case is a sequence of images $x = (x, x_1, x_2,...,x_N)$, such that the first image $x$ is the context image, and the rest are the target images.
    \item 
    For learning without rule induction, all cases are {\it positive cases}, which means that they are all consistent with the background knowledge. For example, in \citepfig{fig:model}, the positive cases are those where Mario indeed walks with the right priority. Furthermore, we introduce the label-limited assumption: ground-truth symbol values are provided for only a very small proportion of cases. 
    \item 
    For learning with rule induction, in addition to positive cases, there is also a set of {\it negative cases} in the dataset. Negative cases are inconsistent with the background knowledge. In the example of \citepfig{fig:model}, a case where Mario walks left, or with up priority, can be a negative case. The label-limited assumption is also adopted in this scenario.
    
\end{itemize}

The setting of symbols, background knowledge, and rules to induce is task-dependent. We refer the detailed setup in our experiments to \citepsec{sec:app:exp}. Assuming both positive and negative cases follows the tradition of inductive logic programming~\citep{muggleton1994inductive,cropper2022inductive}, on which our learning approach is based. The learning processes for both two scenarios are illustrated in \citepfig{fig:learn}, which consists of three parts.

{\noindent\bf Neural model learning}. The learning of the neural model is the same under the two scenarios, which is illustrated in the left part of \citepfig{fig:learn}. In each training iteration, a batch of cases is sampled. The images are processed by the neural model as illustrated in \citepfig{fig:model}, and their corresponding $Y_G, \hat x, z_{G}, z_{sub}$ are generated ($\hat x$ is generated by setting $z_L \leftarrow z_G$, thus it is a {\it reconstruction} of the context image). The model is trained with three kinds of loss signals.
\begin{itemize}
    \item 
    {\bf Symbol grounding loss}. This is the classification error for the symbol value predictions $Y_G$. Since most of the symbol values are not labeled in the dataset, we instead use logical abduction to obtain abduced labels $Y^{abd}_G$ as the ground truth label. The logical abduction process is illustrated in the right part of \citepfig{fig:learn} (see \citepsec{subsec:abd1} and \citepsec{subsec:abd2} for details). The symbol grounding loss is the cross-entropy classification loss calculated between the grounding label prediction of the neural model $P(Y_G)$ and the abduced label $Y^{abd}_G$.
    \item 
    {\bf Reconstruction loss}. This is to minimize the difference between reconstructed images and the true images, which is a common loss for learning generative models. The reconstruction loss controls the generation quality, which is usually calculated as the mean-square-error (MSE) loss w.r.t. the input image $x$ and the reconstructed image $\hat x$. In dSprites, since the images are normalized into binary-valued $\{0, 1\}$ space, we directly utilize logistic loss instead of MSE.
    \item 
    {\bf Disentanglement loss}. This is the feature learning loss for two purposes: 1) Ensure the disentanglement of symbolic and sub-symbolic features, such that the symbol values of the generated images are fully controllable by the symbol grounding module; 2) Ensure the independence of symbolic embeddings, such that a change in one symbol value will not affect other symbols. The details are in \citepsec{subsec:disentanglement}.
\end{itemize}

{\noindent\bf Logical abduction without rule induction}. This learning process is for the scenario of learning without rule induction, which is illustrated in the upper right part of \citepfig{fig:learn}. The target is to identify abduced labels $Y_G^{abd}$ for symbol grounding learning. The basic mechanism is the process of symbol verification. In each step, a sampling method is utilized to generate a candidate set of symbol values for verification. This set of candidate values is then input into the logical reasoning system to check whether any logical rule in the background knowledge is violated. If there is no violation, the verification process is terminated and the current candidate set is taken as the output. It is easy to see that the sampling method plays a crucial role in improving the efficiency of this symbol verification process. The total number of possible symbol groundings is exponential to both the number of symbols per image and the number of images in a case. A good sampling strategy would assign a high probability to promising symbol values, which increases the efficiency of symbol verification. This is also one of the fundamental challenges in abductive learning. We introduce an attention-based symbol sampling strategy to deal with this challenge.

{\noindent\bf Logical abduction with rule induction}. This learning process is for the scenario of learning with rule induction, which is illustrated in the lower right part of \citepfig{fig:learn}. The target is to identify abduced labels $Y_G^{abd}$ and induce new logical rules simultaneously. We introduce a learning method based on the meta-abduction method~\citep{dai2020abductive}, while introducing new techniques to utilize negative cases.

\begin{figure*}[t]
\centering
\includegraphics[width=.9\linewidth]{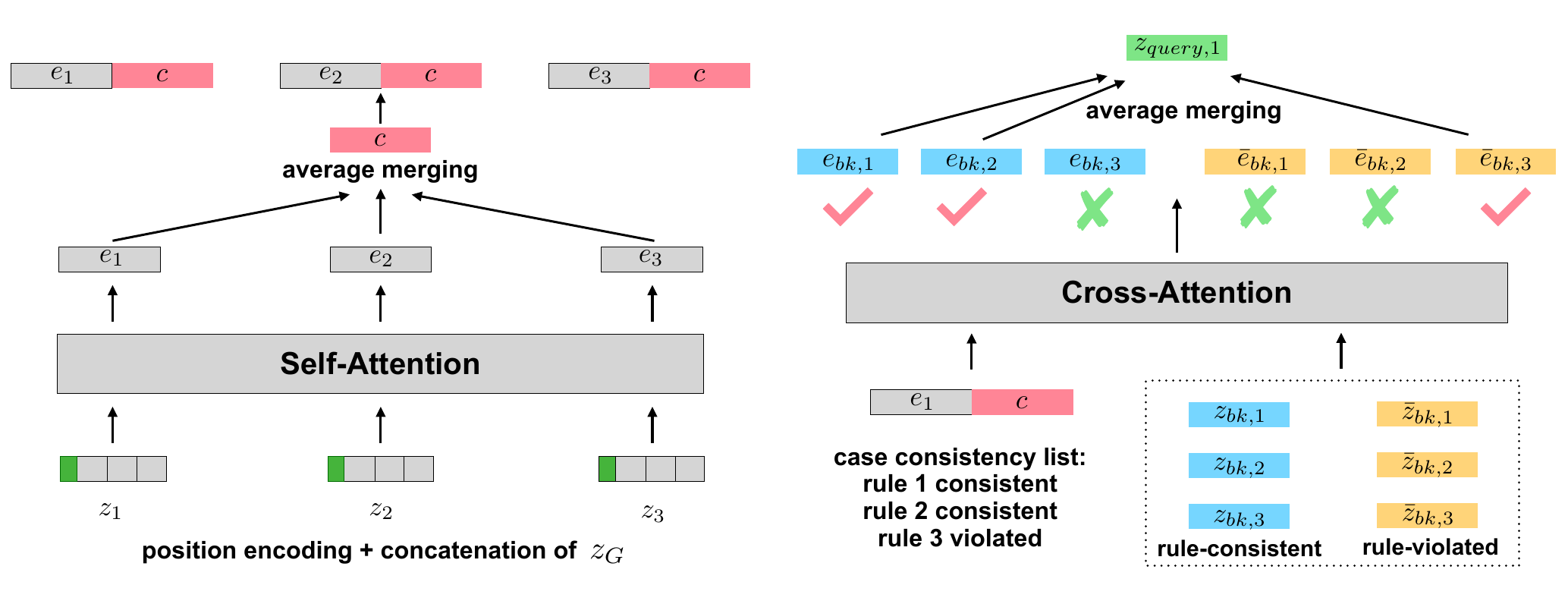}
\caption{The attention-based symbol verification model.} 
\label{fig:verify}
\end{figure*}

\subsection{Attention-Based Symbol Verification}
\label{subsec:abd1}
For the logical abduction process without rule induction as illustrated in the upper right part of \citepfig{fig:learn}, we propose the Attention-based symbol Verification (\vecabd) approach. Toward building a strong symbol grounding sampling strategy, we propose an attention-based model for deciding the sampling probabilities. The model structure is illustrated in \citepfig{fig:verify}. The inputs to the model include 1) the current candidate symbol groundings of a case, as the form of their grounding embeddings; 2) the symbol verification result of the current symbol groundings w.r.t. the background knowledge rules, which can be represented as a binary-valued list, with each value representing whether a rule is violated. The output of the model is the sampling probability for choosing the next candidate symbol grounding, which is calculated by the following process.

{\noindent\bf Self-attention}. The model is illustrated on the left of \citepfig{fig:verify}. For each image in the case, we first obtain their image embeddings $z$ by concatenating all their current candidate grounding embeddings $z_G$, as well as the one-hot positional encoding representing their positions in the case. Then the image embeddings pass the self-attention to obtain after-attention embeddings $e$. We further obtain the embedding $c$, which integrates the information from all images. Finally, we concatenate $c$ with $e$ for each image. By these operations, the final image embeddings integrate sufficient case-level information for further processing.

{\noindent\bf Cross-attention}. The model is illustrated on the right of \citepfig{fig:verify}. The core is a cross-attention module that integrates logical background information with image embeddings. The module has a set of rule embeddings. For each background knowledge rule, there are a pair of rule embeddings, where $z_{bk}$ represents the consistency of the rule, and $\bar z_{bk}$ represents its violation. The cross-attention module takes both image and rule embeddings as inputs and outputs an embedding for each rule with integrated image information. Afterwards, the embeddings are selected based on the input symbol verification result and merged to form the final query embedding $z_{query}$ for the image.

{\noindent\bf Sampling}. The query embedding $z_{query}$ is utilized to obtain their distances to all grounding embeddings $z_{G}$. Finally, we use the softmax transformation of these distances to build the sampling probabilities of symbol groundings for each image. Since the model only involves information from the symbolic part, this model can be well pre-trained using the small part of instances with ground-truth symbol grounding labels in the dataset, by supervising the final sampling probability converging to choose the ground-truth labels. The reason is that symbolic rules are relatively stable and have fewer variations than sub-symbolic information from the data, thus can be well-trained using a much smaller dataset. 

{\noindent \bf Symbol verification}. The symbol verification process is as follows. For one case, the images are processed by the neural model and the neural predicted symbol grounding values $Y_G$ are generated. Afterwards, the values are sent to the logical reasoning system to verify their consistency with all the rules in the background knowledge $B$, and then the case consistency list is generated. Subsequently, symbol verification model processes the information, and generates $z_{query}$ as illustrated in \citepfig{fig:verify}. Then $z_{query}$ is treated as $z_{sym}$, and a label sampling distribution $P_V(Y)$ is calculated based on \citepeq{eq:metric}. The next step is to sample a fixed-sized pool of candidate symbol values from $P_V(Y)$, and check their consistency with $B$ in sequence. If none of the candidates pass the consistency test, the case is discarded for symbol grounding learning. 
The training process consists of the following two stages. 

{\noindent\bf Pre-training}. At this stage, both the neural generation model and the symbol verification model are pre-trained with the small labeled subset of the training data. The generation model is updated using ground-truth images and symbol value labels. The symbol verification model is updated from the cross-entropy loss calculated between $P_V(Y)$ and the ground-truth symbol value labels.

{\noindent\bf Training}. At this stage, the learning without rule induction process is conducted using the whole training dataset. Note that the symbol verification model is also updated during this process, where the ground-truth labels in the pre-training stage are substituted with the abduced labels $Y^{abd}_G$. While the model is only updated by those instances that the initial symbol grounding $Y_G$ is not consistent with the abduced label $Y^{abd}_G$. This can avoid the catastrophic forgetting of the pre-training experience for the symbol verification model. 

\subsection{Contrastive Meta-Abduction}
\label{subsec:abd2}

For the logical abduction process with rule induction, which is illustrated in the lower right part of \citepfig{fig:learn}, we propose the Contrastive Meta-Abduction (\conabd) approach. To conduct the abduction of grounding labels and induction of hidden rules, the system tries to solve the following learning problem:
\begin{equation}
  (Y^{abd}_G, H) = \argmax_{Y, H} P(Y, H|B, x),
  \label{eq:metaabd1}
\end{equation}
in which $H$ is the hidden rule and $x$ represents the cases of images. The equation shows that the target is to find the grounding label and hidden rule with the maximum posterior probability. The posterior probability can be further decomposed as
\begin{equation}
  (Y, H|B, x) = P(H|B, Y)P(Y|x).
  \label{eq:metaabd2}
\end{equation}
This decomposition reveals the basic mechanism of meta-abduction. $P(Y|x)$ is set as $P(Y_G)$, which is generated by the neural model. $P(H|B, Y)$ is generated by the meta-abduction program implemented in logical background knowledge, which is calculated by 
\begin{equation*}
  \begin{cases}
    P(H|B, Y) \propto  1/ [\mathrm{length}(H)]^2, \\
    \quad \text{if}\; B\cup H \models Y_P, B\cup H \not\models Y_N, \\
    P(H|B, Y) = 0, \text{otherwise},
  \end{cases}
  \label{eq:metaabd3}
\end{equation*}
in which $\models, \not\models$ denote logically entails and cannot entail respectively. In this equation, we have $Y=Y_P\cup Y_N$, such that $Y_P$ denotes all symbol groundings in the positive cases, $Y_N$ denotes symbol groundings in the negative cases. Furthermore, $\mathrm{length}(H)$ denotes the number of clauses in $H$. This means that the meta-abduction program considers all $H$ that logically entails $Y_P$ given $B$, meanwhile not entails $Y_N$ given $B$, and prefers simpler rules with fewer clauses. To search for a possible $H$, the meta-interpretative learning method Metagol~\citep{cropper2020learning} is utilized. The basic mechanism of Metagol is to learn first-order logical programs with a second-order meta-interpreter based on first-order background knowledge $B$ and a set of {\it meta-rules}, which defines the possible rule space to search. The design of meta-rules is a crucial task, and we report our design on each dataset in the experimental setup section (\citepsec{sec:app:exp}).

Note that for inducing the precise hidden rule, it is ideal to achieve the following condition: Under the ground truth grounding label $Y^*$, the ground-truth hidden rule $H^*$ satisfies $P(H^*|B, Y^*) \approx 1$, such that there is no other $H$ logically entailing $Y^*$. This condition is quite strong for complicated hidden rules, which is difficult to satisfy when only positive cases are used for learning. However, the existing meta-abduction method MetaAbd~\citep{dai2020abductive}, which builds a mechanism similar to above, does not support using negative cases in learning. The reason can be seen from \citepeq{eq:metaabd1} and \citepeq{eq:metaabd2}. The abduction solution is chosen from all possible symbol groundings. For negative cases, it is quite likely that trivial symbol groundings exist to make the not entailling constraint meaningless. For example, setting Mario remains in a fixed position for all images is such a trivial grounding, such that the not-entailling constraint indeed satisfies if we do not allow such standing still situation according to the background knowledge. To avoid trivial grounding, limiting the choice of negative case symbol groundings is essential. However, for MetaAbd, such a limitation is difficult to realize.

\noindent{\bf Symbol binding}. The vector-quantized symbol grounding mechanism for our model provides a natural and easy way for limiting negative case symbol groundings. We introduce symbol binding as a preprocessing process before the abduction process in \citepeq{eq:metaabd1} starts. For some symbol to ground in the negative case, we first identify its most likely value $Y$ according to $P(Y_G)$. Then we fetch the grounding embedding $z$ from the dictionary of the symbol grounding module. The same operation is also performed for all corresponding positive case symbols, and the set of positive case embeddings $Z_P$ is obtained. Then we identify the nearest neighbor of $z$ in $Z_P$ by
\begin{equation}
  z_P = \argmin_{z'\in Z_P} \| z-z'\|.
  \label{eq:metaabd4}
\end{equation}
Finally, in the rest of learning process, we introduce an additional constraint that the symbol grounding of $z$ and $z_P$ must be the same. In this way, the trivial solutions for negative case symbol groundings are eliminated.

\subsection{Enabling Symbolic Control on Generation}
\label{subsec:disentanglement}
As discussed in the neural model learning part, the disentanglement loss is crucial to make the neural generation model controllable by symbol guidance. We propose the following two disentanglement losses. 

{\noindent\bf Sub-symbolic adversarial loss $L_{ADV}$.}
To ensure that the symbolic information only passes the symbolic route, we need to force the sub-symbolic embeddings $z_{sub}$ to have as little symbolic information as possible. To achieve this, we build an additional sub-symbolic adversarial classifier, which takes $z_{sub}$ as the input. We make the classifier predict the abduced symbol grounding labels, which are trained with cross-entropy loss. We implement a gradient-inverse layer inspired by~\citet{ganin2016domain}, in between $z_{sub}$ and the classifier, such that $z_{sub}$ is trained to decrease prediction accuracy. The gradient-inverse layer makes the sign of backward-propagated gradient signal inverse:
\begin{equation}
  \nabla_{inv}(z_{sub}) = -\alpha \nabla(z_{sub}),
  \label{eq:ginvlayer}
\end{equation}
in which $\nabla(z_{sub})$ and $\nabla_{inv}(z_{sub})$ denote the original and inverse gradients passed through $z_{sub}$. It can be seen that since the gradient is inverse, the sub-symbolic encoder will try to maximize the prediction loss instead of minimizing it. The hyperparameter $\alpha$ adjusts the strength for adversarial training. In this way, $z_{sub}$ cannot encode symbolic information, since any useful symbolic information in $z_{sub}$ would lead to greater accuracy of the adversarial classifier prediction.

{\noindent\bf Symbolic GAN loss $L_{GAN}$.} We further introduce a GAN loss to enhance the feature independence among grounding embeddings $z_G$, which is motivated from the face age regression studies~\citep{zhifei2017cvpr,zhu2018facial}. The main idea is to randomly generate symbol values for training images and substitute the original grounding embeddings with the corresponding embeddings to generate the target image with the random symbol values. Furthermore, a set of conditional discriminators is introduced to discriminate the generated image from the real image with the same symbol grounding value in the dataset. Note that the real images are actually unlabeled, thus we utilize their abduced labels as ground-truth labels instead. The GAN loss is introduced to make the neural model the adversary of the discriminator. The intuition is that the generation results for randomly chosen symbol values can only be correct when the independence among symbols is maintained.

\section{Experiments}
\label{sec:exp}

\begin{figure*}[t]
    \centering
    \begin{subfigure}[t]{.24\textwidth}
        \includegraphics[width=\textwidth]{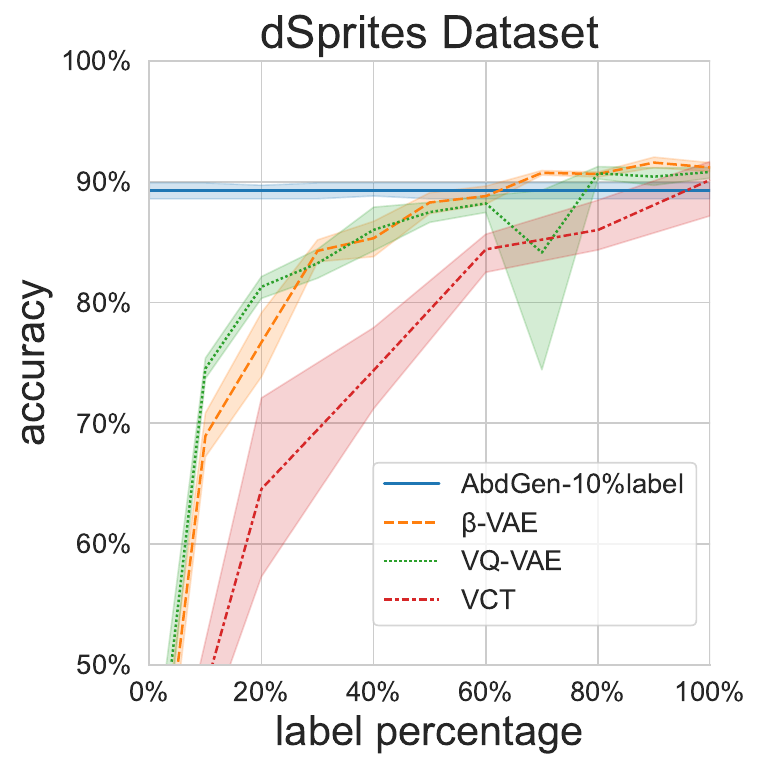}
    \end{subfigure}
    \begin{subfigure}[t]{.24\textwidth}
        \includegraphics[width=\textwidth]{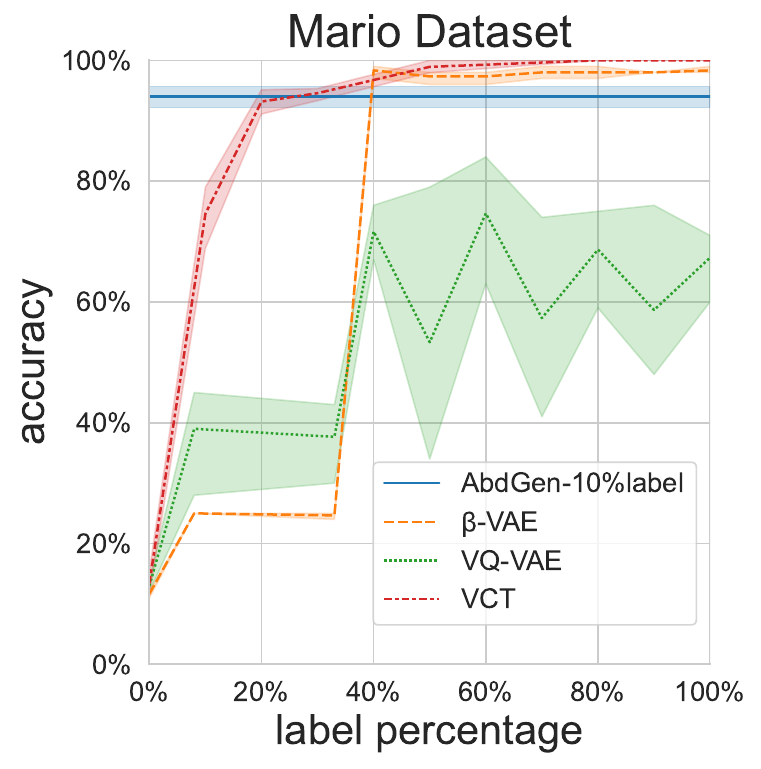}
    \end{subfigure}
    \begin{subfigure}[t]{.228\textwidth}
        \includegraphics[width=\textwidth]{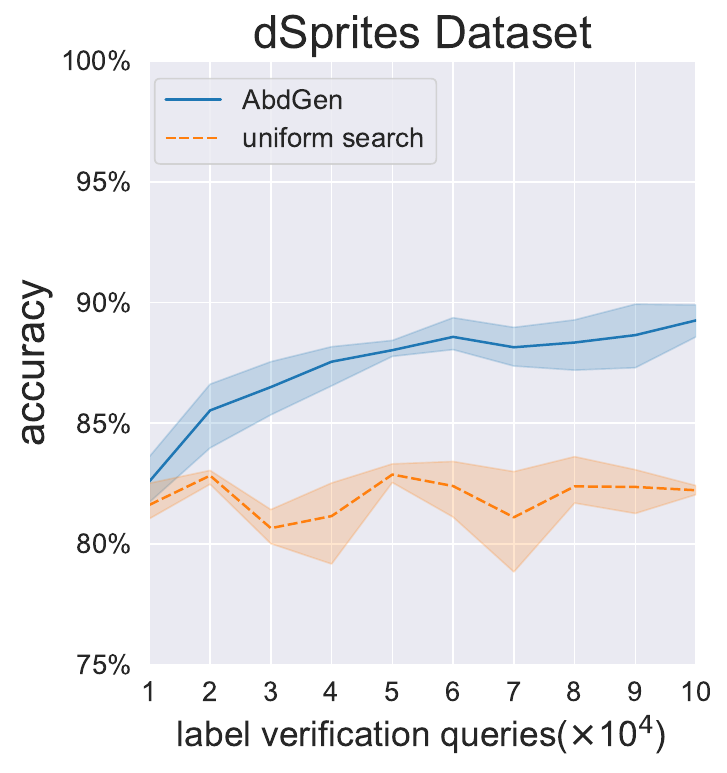}
    \end{subfigure}
    \begin{subfigure}[t]{.235\textwidth}
        \includegraphics[width=\textwidth]{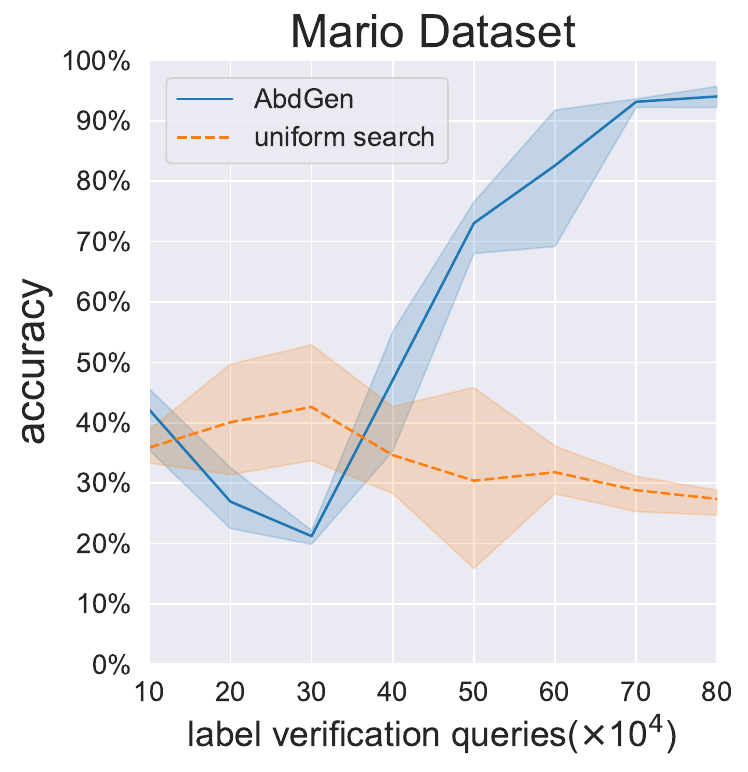}
    \end{subfigure}
    \caption{{\bf Learning with rule induction}. The two subfigures on the left illustrate the grounding accuracy ($\uparrow$). The two subfigures on the right demonstrate the accuracy w.r.t. number of label verification queries for AbdGen and uniform sampling. The means and standard errors are obtained from three random repeats.}
    \label{fig:grounding}
\end{figure*}

\begin{figure*}[t]
    \centering
    \begin{subfigure}[t]{.23\textwidth}
        \includegraphics[width=\textwidth]{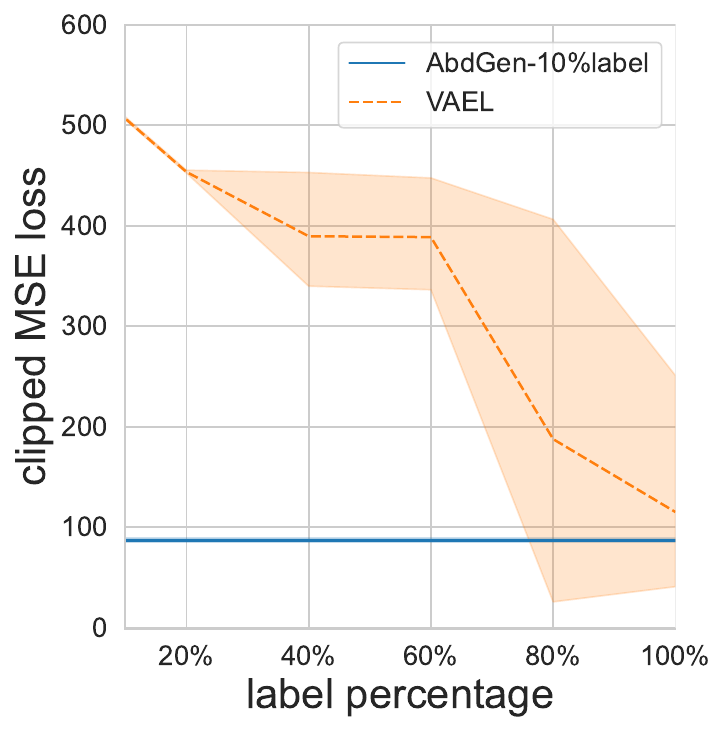}
    \end{subfigure}
    \begin{subfigure}[t]{.24\textwidth}
        \includegraphics[width=\textwidth]{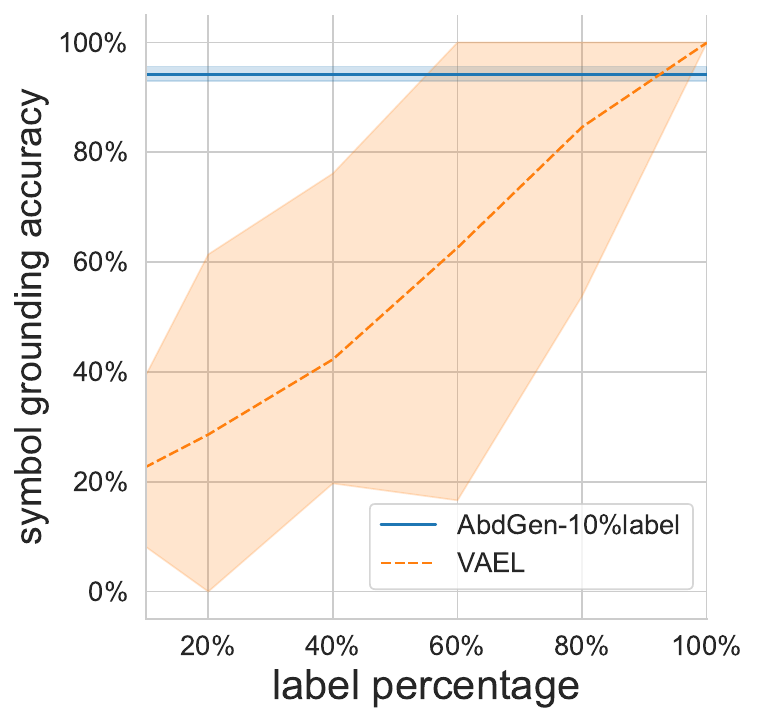}
    \end{subfigure}
    \begin{subfigure}[t]{.23\textwidth}
        \includegraphics[width=\textwidth]{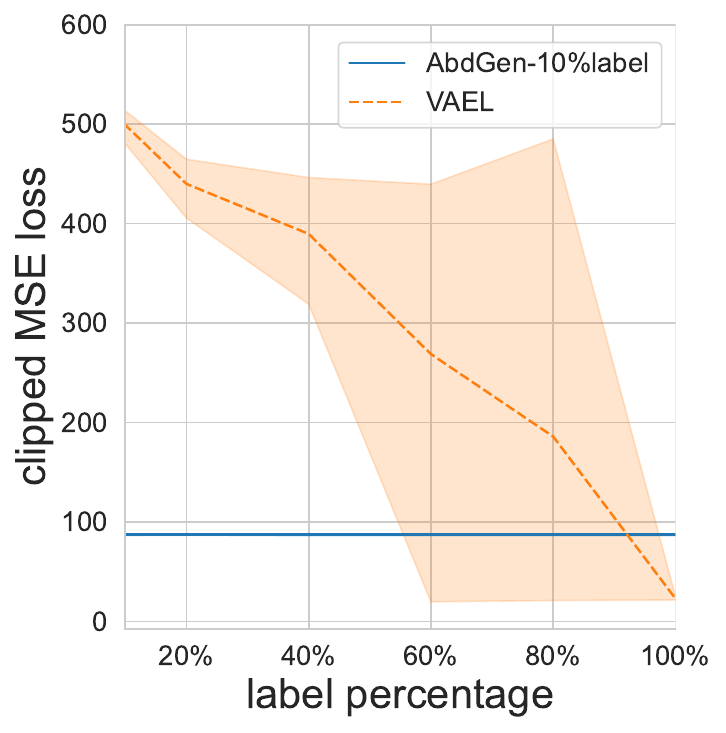}
    \end{subfigure}
    \begin{subfigure}[t]{.24\textwidth}
        \includegraphics[width=\textwidth]{Images/mario_label_sym_acc_ld.pdf}
    \end{subfigure}
    \caption{{\bf Generation under unseen rules}. Comparisons of AbdGen and VAEL on clipped MSE loss ($\downarrow$) and grounding accuracy ($\uparrow$). The first two subfigures correspond to the right-up priority configuration while the other two correspond to the left-down priority configuration.}
    \label{fig:unseenrule}
\end{figure*}

We verify the following questions from experiments: 1) whether AbdGen requires significantly fewer labels for symbol grounding compared to the baselines; 2) Compared to VAEL, whether AbdGen can achieve similar or stronger ability in generalizing to unseen rules; 3) whether AbdGen can effectively conduct rule induction. The detailed experimental setups are introduced in \citepsec{sec:app:exp}.

\subsection{Learning without Rule Induction}

 The experiments are carried out on dSprites~\citep{dsprites17} and Mario~\citep{misino2022vael} Datasets. Three baseline approaches, VQ-VAE~\citep{van2017neural}, $\beta$-VAE~\citep{higgins2016beta}, and VCT~\citep{yang2022visual} are introduced. The baselines are chosen since they are typical approaches with strong disentanglement abilities, which would be helpful under the limited label scenario. If AbdGen could outperform these baselines, the effectiveness of logical weak supervision would be justified. To evaluate the grounding accuracy under limited instance-level information, we train all the baselines under varying percentages of labeled data, except for AbdGen, on which we only utilize 10$\%$ of training data to pre-train the \vecabd\;model. For all baselines, we augment them with single-layer linear classifiers upon their embedding layers for symbol grounding.

{\bf Results}: Performance comparison is shown in the two subfigures on the left of \citepfig{fig:grounding}. With only 10$\%$ labeled data, AbdGen can approach the baselines when using fully labeled data. This solidly verifies the effectiveness of symbol grounding based on logical supervision. Furthermore, the two right subfigures of \citepfig{fig:grounding} illustrate the convergence speed during training with respect to the number of symbol label verification calls to the logical reasoning system. It can be seen that AbdGen is much more efficient than the uniform sampling strategy, verifying the effectiveness of \vecabd. Furthermore, \citeptab{tab:add_ablation1} shows the grounding accuracy of AbdGen under different percentages of labeled data. It
shows that AbdGen achieves even better performance given sufficient labeled data. 

\begin{figure}[t]
    \centering
    \begin{subfigure}[t]{.5\linewidth}
        \includegraphics[width=\linewidth]{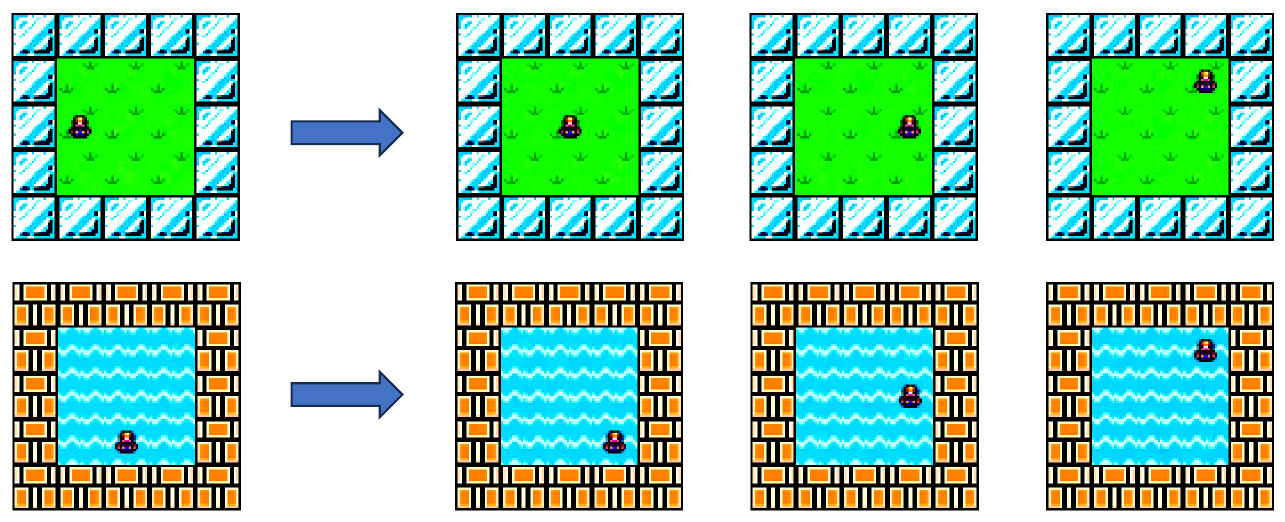}
    \end{subfigure}
    \begin{subfigure}[t]{.5\linewidth}
        \includegraphics[width=\linewidth]{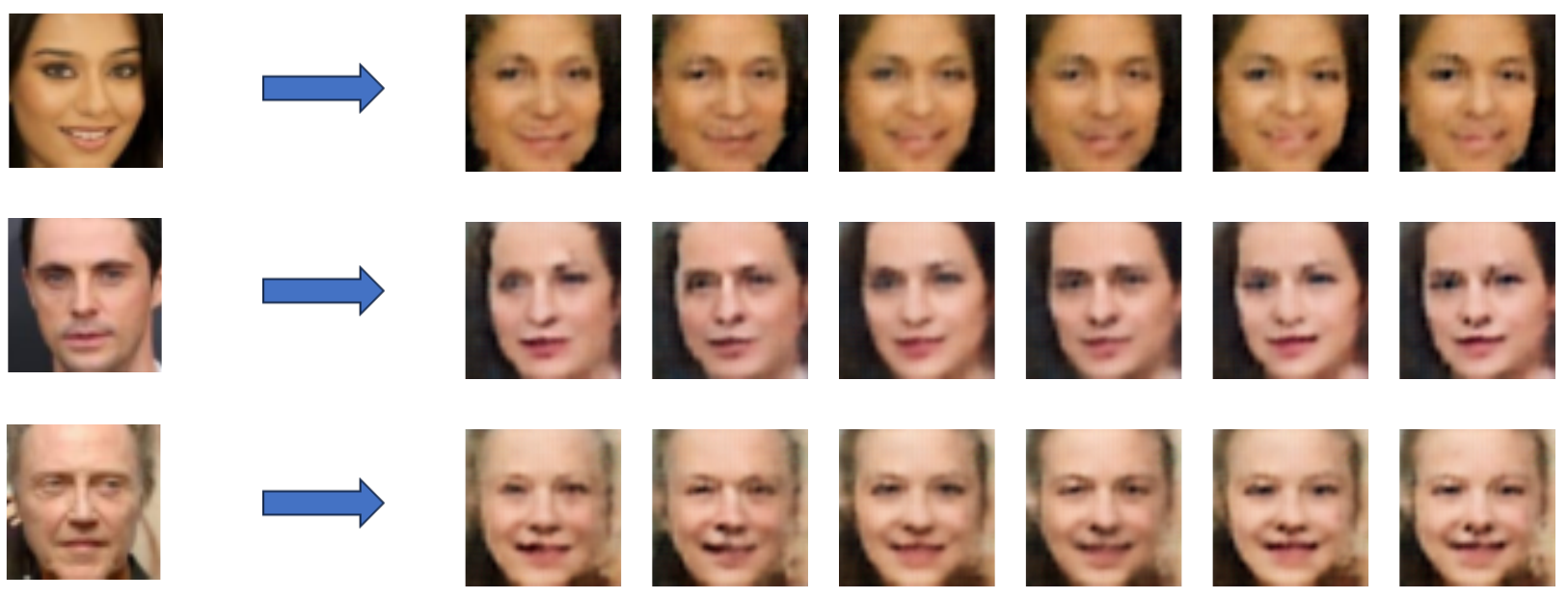}
    \end{subfigure}
    \caption{{\bf Learning with rule induction}. Generation results of \method\;on Mario (above) and UTKFace (below). Given the context image, target images are generated based on the learned rule. More results are reported in \citepfig{fig:moreface}.}
    \label{fig:rulelearn}

\end{figure}

\noindent{\bf Generalization to Unseen Rules}. This task is still under the scenario of learning without rule induction. Our objective is to compare AbdGen with VAEL~\citep{misino2022vael} in the task of image generation based on the testing rule that is different from training. Following the setting in the VAEL paper, we conduct the experiments under the Mario dataset. We consider two configurations: 1) the training rule is going with right priority first and then with up priority, while the testing rule is going one-step down or left; 2) the training rule is going with left priority first and then with down priority, while the testing rule is going one-step up or right. Following the VAEL paper, we provide VAEL with labels for pairs of images in the cases, which provide the relative moving direction from one to another. For AbdGen, we still keep the setting of using 10\% of training labels for pre-training the \vecabd\;model.

The results are shown in \citepfig{fig:unseenrule}. We report the comparison of two performance measures: 1) clipped MSE loss, which is calculated by clipping the original and generated images into 20$\times$20 pieces, and choosing the largest MSE loss among all pieces. This would make the results more significant since the images in the Mario dataset usually have similar and duplicated backgrounds; 2) symbol grounding accuracy. The results show that AbdGen requires significantly less labeling information to achieve a similar performance of VAEL trained under full labels. We also provide a qualitative comparison of generated images in \citepfig{fig:exp_newrulegen}.

\subsection{Learning with Rule Induction}
We evaluate the rule induction ability of AbdGen under the Mario and UTKFace~\citep{zhifei2017cvpr} datasets. UTKFace includes facial images annotated with ages and genders. We design the underlying generation rule (correct rules in Tab. 1) on the Mario dataset as {\it moves with right priority if possible, and then move up}. Under UTKFace, we design the rule as generates with {\it age descending priority, and with female gender priority when the ages equal}. Our testing procedure involves presenting an image from the test set and then generating subsequent images based on acquired rules. We compare AbdGen with MetaAbd~\citep{dai2020abductive}. We apply the AbdGen visual generator to MetaAbd, making MetaAbd capable of performing visual generation.

{\bf Results}: \citeptab{tab:rule} illustrates examples of the logical rules under the two datasets, which include wrong, less informative, and the ground-truth correct rules. From \citeptab{tab:exp3_perc}, it can be observed that AbdGen significantly outperforms MetaAbd in learning correct rules. Without making use of information from the negative cases, MetaAbd fails to distinguish between correct and incorrect rules, as well as eliminate less informative rules. This phenomenon verifies the importance of including contrastive information from both positive and negative cases for rule abduction. The qualitative generation results of the trained AbdGen model are illustrated in \citepfig{fig:rulelearn} and \citepfig{fig:moreface}. Given input images, the model generates a sequence of images following the learned generative rules. In particular, under UTKFace, the model captures complex semantic groundings of ages and genders. 

Furthermore, \citeptab{tab:add_ablation2} shows ablations on disentanglement losses under learning with rule induction. Under Mario, ground-truths are all
known, so the clipped MSE is reported. Under UTKFace, ground-truths for changed images are unknown, so we report their $L_2$
distances w.r.t. the input image
(larger the better, since it indicates variety). The results clearly indicate the essentialness of both disentanglement losses.

\begin{table}[t]
\centering
\caption{Grounding accuracy of AbdGen under different percentages of labeled data for learning without rule induction.
}
\label{tab:add_ablation1}
\resizebox{.5\linewidth}{!}{
\scriptsize

\begin{tabular}{ccccc}
\toprule
Dataset & 5\% & 10\% & 50\% & 100\% \\
\midrule
dSprites  & 84.7\%   & 89.3\%  & 94.2\%  & 97.0\%     \\ 
Mario & 86.9\% & 94.0\% & 99.5\% & 100.0\% \\ 
\bottomrule
\end{tabular}}
\end{table}

\begin{table}[t]
\centering
\caption{Ablations on disentanglement losses under learning with rule induction: w/o adv. indicates without subsymbolic adversarial loss, and w/o GAN indicates without symbolic GAN loss.
}
\label{tab:add_ablation2}
\resizebox{.5\linewidth}{!}{
\scriptsize

\begin{tabular}{cccc}
\toprule
Dataset (Metric) & Ours & w/o adv. & w/o GAN \\
\midrule
Mario (Clip. MSE$\downarrow$) & 27.8 & 31.4  & Not Used     \\ 
UTKFace (Gen. Diff.$\uparrow$) & 2900.9 & 508.3 & 514.7 \\ 
\bottomrule
\end{tabular}}
\end{table}

\subsection{Text-to-Image Generation}

Our method can also be applied to other generation tasks, such as text-to-image generation. Here, we utilize the pre-trained diffusion model proposed in Attend-and-Excite~\citep{chefer2023attend} to conduct symbol grounding, rule induction, and logic-guided generation by adapting AbdGen into its generation pipeline. We use the attention map during diffusion to ground unstructured images and learn the rules by ConMetaAbd. The new prompts following the rule are then delivered to Attend-and-Excite to generate new images. The neural part of this approach does not require a training process.

\noindent{\bf Grounding}. To induce rules from unstructured images, the images need to be grounded into multiple symbolic subjects. Instead of using a specific grounder, we compared the maximum values of cross-attention maps introduced in Attend-and-Excite, given prompts such as ``a photo of a dog'' related to different objects, to roughly ground the images into $k$ symbols. This grounding process is performed under the assumption that the prompts related to objects in images show relatively large maximum values of cross-attention maps. However, the maximum value is not a precise degree of multi-label classification. Thus, the grounding results are often noisy. Fortunately, MetaAbd can still learn the correct rules if the top $k$ grounding symbols contain the rule-related ones.

\noindent{\bf Symbol binding}. The negative samples also bind to the positive samples following \citepeq{eq:metaabd4}. In this case, $z$ is the maximum value of the cross-attention map of each sample, which means that we identify the nearest positive neighbor image of the negative image according to the semantic similarity. For example, an image containing ``dog'' and ``bone'' is more similar to an image with ``dog'' and ``hat'' than it is with ``cat'' and ``fish''.

\begin{table*}[t]
\centering
\caption{Illustration of examples of the induced rules in the learning with rule induction task, providing examples of correct, incorrect, and less-informative rules for both Mario and UTKFace.   
}
\label{tab:rule}
\resizebox{.9\linewidth}{!}{
\begin{tabular}{cccc}
\toprule
\multirow{2}{*}{Dataset} & & Rule Example & \\
& Wrong & Less-Informative & Correct \\ 
\midrule
Mario & 
\makecell{$\mathtt{f(A,B)}$:-$\mathtt{down(A,C),f(C,B)}.$ \\
$\mathtt{f(A,B)}$:-$\mathtt{f_1(A,B)}$.\\
$\mathtt{f_1(A,B)}$:-$\mathtt{left(A,C),f_1(C,B)}$.\\
$\mathtt{f_1(A,B)}$:-$\mathtt{terminate(A,B)}$.\\} 
& \makecell{$\mathtt{f(A,B)}$:-$\mathtt{right(A,C),f(C,B)}$. \\
$\mathtt{f(A,B)}$:-$\mathtt{up(A,C),f(C,B)}$.\\
$\mathtt{f(A,B)}$:-$\mathtt{terminate(A,B)}$.\\} 
& \makecell{$\mathtt{f(A,B)}$:-$\mathtt{right(A,C),f(C,B)}$. \\
$\mathtt{f(A,B)}$:-$\mathtt{f_1(A,B)}$.\\
$\mathtt{f_1(A,B)}$:-$\mathtt{up(A,C),f_1(C,B)}$.\\
$\mathtt{f_1(A,B)}$:-$\mathtt{terminate(A,B)}$.}  \\ 
\midrule
UTKFace
& \makecell{$\mathtt{f(A)}$:-$\mathtt{gender(A,B),greater}$\_$\mathtt{than(B)}$. \\
$\mathtt{f(A)}$:-$\mathtt{gender(A,B),equal(B),ages(A,C),greater}$\_$\mathtt{than(C)}$. \\} 
& \makecell{$\mathtt{f(A)}$:-$\mathtt{ages(A,B),greater}$\_$\mathtt{than(B)}$. \\
$\mathtt{f(A)}$:-$\mathtt{gender(A,C),great}$\_$\mathtt{than(C)}$. \\} 
& \makecell{$\mathtt{f(A)}$:-$\mathtt{ages(A,B),greater}$\_$\mathtt{than(B)}$. \\
$\mathtt{f(A)}$:-$\mathtt{ages(A,B),equal(B),gender(A,C),greater}$\_$\mathtt{than(C)}$. \\ } \\
\bottomrule
\end{tabular}}

\end{table*}

\begin{figure*}[t]
\centering
\includegraphics[width=0.9\textwidth]{Images/text2image_data_generate.pdf}
\caption{Text-to-image generation results. Each row illustrates an individual generation task. After logical abduction based on the negative and positive training examples, the model generates sequences of images following the guiding prompts and the induced rules.} 
\label{fig:text2image}
\end{figure*}

We use DALL-E 3~\citep{dalle3} to generate four sets of examples corresponding to four rules. Each set includes 20-30 images. For each set, we randomly sample to formulate 100 training instances, each including two positive cases and six negative cases. A positive case is an image sequence with rule-related subjects in the correct order following the rule, while a negative case breaks the rule. 
We define a dictionary of subjects, as well as a predicate set in the logical background knowledge. Each predicate specifies one ranking among a subset of subjects. After successful rule induction, a predicate is identified. The learned rules are then used to choose the subjects according to the ranking defined by the chosen predicate for generating a sequence of images. The results are illustrated in \citepfig{fig:text2image} and \citeptab{tab:exp3_perc}, showing that our approach can be successfully integrated with pre-trained text-to-image models.

\begin{table}[t]
\centering
\caption{Comparison on the proportions of correct, incorrect, and less informative rules produced by AbdGen and MetaAbd under Mario and UTKFace datasets, as well in the text-to-image generation task.
}
\label{tab:exp3_perc}
\resizebox{.5\linewidth}{!}{
\scriptsize

\begin{tabular}{ccccc}
\toprule
\multirow{2}{*}{Dataset} & \multirow{2}{*}{Method} &  & Proportions of Induced Rules& \\
& & Wrong & Less-Informative & Correct \\ 
\midrule
\multirow{2}{*}{Mario}  
& MetaAbd      & 0.11     & 0.89     & 0     \\ 
& AbdGen       & 0.05 & 0.12 & 0.83 \\ 
\midrule
\multirow{2}{*}{UTKFace} 
& MetaAbd      & 1     &0      & 0      \\ 
& AbdGen       & 0.03 & 0 & 0.97 \\ 
\midrule
\multirow{2}{*}{Text-to-Image} 
& MetaAbd      & 0.52     & -      & 0.48      \\ 
& AbdGen       & 0.18 &  - & 0.82 \\ 
\bottomrule
\end{tabular}}

\end{table}

\section{Conclusion}
In this paper, we propose the AbdGen approach in bridging neural visual generation and logical reasoning. We propose a logic-integrated neural visual generation model and the corresponding symbol grounding learning method, leading to the effective disentanglement of symbolic and sub-symbolic information and controllable symbol-based generation. An attention-based symbol verification mechanism is proposed to boost the efficiency of symbol grounding learning. Furthermore, AbdGen is capable of doing rule induction based on contrastive meta-abduction. Experimental results verify that our approach effectively utilizes logical supervision for symbol grounding learning instead of labeling information and, meanwhile, is capable of performing precise rule induction.

{\noindent \bf Limitations and future work}. 
Aiming at proposing an initial neurosymbolic generative learning framework, we focus on tasks which are sufficient to justify how well our approach addresses the core technical challenges without requiring complicated neural and symbolic sub-modules. Future research can indeed utilize our framework to tackle more challenging tasks by utilizing stronger sub-modules.
Furthermore, a major challenge of conducting visual generative abductive learning is to determine the logical background knowledge without significant human-designed priors. Integrating the reasoning system with language models and external knowledge bases can be a promising future solution.

\section*{Acknowledgment}
This work was supported by National Key R\&D Program of China (2022ZD0114804), National Natural Science Foundation of China (62206245, 62206124), Jiangsu Science Foundation
Leading-edge Technology Program (BK20232003), and Major Program (JD) of Hubei Province (2023BAA024). We would like to thank the anonymous reviewers for their constructive suggestions.

\bibliographystyle{named}
\bibliography{nesy}

\clearpage
\appendix

\section{Experimental Setup}
\label{sec:app:exp}
The neural and logical parts of AbdGen are implemented with PyTorch~\citep{paszke2019pytorch} and SWI-Prolog~\citep{wielemaker2003overview} respectively. The experiments are conducted on a server with AMD Ryzen Threadripper Pro 3975WX processor, 256GB RAM, and four NVIDIA RTX 4090 GPUs. 
\subsection{Dataset Description}
\begin{itemize}
\item
{\bf dSprites~\citep{dsprites17}.}
The dSprites dataset comprises 132,710 images used for reconstruction pre-training. The test set includes 14,745 images with dimensions 64 $\times$ 64 $\times$ 1. These images are characterized by shape, scale, orientation, and $x$ and $y$ axis-positions. For simplicity, we excluded the $x,y$ positions and discretized the orientation into 4 quadrants each spanning 90 degrees. The dataset identifies 3 distinct shapes and 6 different scales.

\item
{\bf Mario~\citep{misino2022vael}.}
The Mario dataset provides a total of 12,600 images for reconstruction pre-training and unlabeled weakly supervised training. Each image of size 100 $\times$ 100 $\times$ 3 displays Mario's position on a 3 $\times$ 3 grid. In this dataset, only Mario's position varies across images, with the surrounding environment remaining constant.

\item
{\bf UTKFace~\citep{zhifei2017cvpr}.} The UTKFace dataset consists of 24107 images with age and gender annotations. To conduct rule learning, we randomly sampled 20301 images for training, which are cropped and resized into 128 $\times$ 128 $\times$ 3. The test set comprises 2237 images of the same size. We utilize the original genders of "male and female" annotations of the images as ground-truth, meanwhile build three age categories into "young (less than 18), middle (between 18 and 60), and old (beyond 60)" based on the concrete ages. We constructed 10000 random sets as training cases, each containing 6 images representing different gender-age identities.
\end{itemize}

\subsection{Learning Without Rule Induction}
\begin{itemize}
  \item  
    {\bf dSprites}
    \begin{itemize}
      \item  
        {\bf Overview.}
        We choose three symbols, shape, scale, and orientation, out of all latent factors in the original dSprites dataset, as the grounding symbols in the experiment. We simplify the three symbols into three-class, three-class, and four-class symbol values, respectively. We design specific logical rules as the logical background knowledge to generate cases of pairs of images.
      \item
        {\bf Logical background knowledge.} Four logical rules are introduced: \begin{enumerate}
          \item 
            The class of the scale of the first image is smaller than the second for -1.
          \item 
            The class of the orientation of the first image is smaller than the second for -1.
          \item
            The shape has the same label as the scale in each image.
          \item 
            For each image, the shape with class 0 can only have orientation 0 or 1, the shape with class 1 can only have orientation 1 or 2, the shape with class 2 can only have orientation 2 or 3.
        \end{enumerate}
      \item
        {\bf Data preparation.} We sample 2000 images from the original dataset, and construct
10000 training image cases based on the logical background knowledge rules. The reported assignment accuracy is evaluated on holdout testing images.
      \item
        {\bf Experimental setup.} For baseline methods, the training process is conducted for 15000 iterations. For AbdGen, the learning is conducted through 100000 symbol verification calls with 10000 iterations of pre-training for the attention-based symbol verification module.
    \end{itemize}
  \item 
      {\bf Mario}
      \begin{itemize}
        \item 
          {\bf Overview.} We choose the position of Mario as the semantic factor, which has nine possible values. We design specific logical rules as background knowledge to generate groups of five images for training. The target is to learn neural generators to generate the moving step trajectories of Mario by successive images.
        \item
          {\bf Logical background knowledge.} We include the specific logical rule that Mario first moves with the right priority if possible, then with up priority. 
        \item
          {\bf Data preparation.} 
          The dataset is constructed by cases of images.
          Each case contains three to five images, forming a complete trajectory of the agent's movement. The overall training set includes 4536 images, and the testing set includes 504 images.
        \item
          {\bf Experimental setup.} For baseline methods, the training process is conducted for 10000 iterations. For AbdGen, the learning is conducted through 800000 symbol verification calls with 10000 iterations of pre-training for the attention-based symbol verification module. 
\end{itemize}
\end{itemize}

\subsection{Learning with Rule Induction} 
\begin{itemize}
  \item 
    {\bf Mario}
    \begin{itemize}
      \item
        {\bf Logical background knowledge.} 
            \begin{enumerate}
              \item The system is given four basic movement functions, defining moving towards each direction for one-step.
              \item Integrity constraint: defining that the Mario can only be at one position in one image.
              \item 
              The terminate positions of all cases are given.
            \end{enumerate}
      \item 
      {\bf Latent rule to induce}. Mario moves with right priority if possible, and then move up.
      \item
        {\bf Meta-rules.} Meta-rules are fundamental ingredients of meta-interpretative/abductive learning, which are second-order logical rule templates for rule learning. We utilize three meta-rules:
        $$\mathtt{metarule(ident, [P,Q], [P,A,B], [[Q,A,B]])}$$
        $$\mathtt{metarule(chain, [P,Q,R], [P,A,B], [[Q,A,C],[R,C,B]])}$$
        $$\mathtt{metarule(tailrec, [P,Q], [P,A,B], [[Q,A,C],[P,C,B]])}$$
      \item
        {\bf Data preparation.} The basic dataset setup is similar to symbol assignment experiments. For rule learning, we construct 672 positive cases and 2520 negative cases using the training data.
      \item
        {\bf Experimental setup.}
        For all methods, the training process is conducted for 20000 iterations. For each iteration, we sample 10 positive cases and 20 negative cases to form an abductive instance. 
    \end{itemize}
  \item 
      {\bf UTKFace}
      \begin{itemize}
        \item 
          {\bf Symbols.}
          We choose two symbols, age, and gender, to be grounded in the experiments. The two symbols have three (young, middle, aged) and two (female and male) possible values, respectively. 
        \item
          {\bf Logical background knowledge.} 
            \begin{enumerate}
              \item The system is given three comparison functions, defining greater than, less than, and equal relationships for the two factors (we assume that the gender is also valued, such that female is 1 and male is 0).
              \item The system is given integrity constraints, defining that each person should have one age and one gender.
            \end{enumerate}

        \item 
          {\bf Latent rule to induce}. Generating with age descending priority, and with female gender priority when the ages equal.
        \item
          {\bf Meta-rules.} 
        We utilize four meta-rules:
        $$\mathtt{metarule([P,Q,R,S,T],[P,A],}$$
        $$\mathtt{[[Q,A,B],[R,B],[S,A,D],[T,D]])}$$
        $$\mathtt{metarule([P,Q], [P,A], [[Q,A]])}$$
        $$\mathtt{metarule([P,Q,R], [P,A], [[Q,A,B],[R,B]])}$$
        $$\mathtt{metarule([P,Q,R], [P,A,B], [[Q,A,C],[R,C,B]])}$$
        \item
          {\bf Data preparation.} We utilize 20301 images for training as described previously. We further randomly construct 10000 groups of six images following the generative rules of positive cases using the training images. 

        \item
          {\bf Experimental setup.} For all methods, the training process is conducted for 100000 iterations. We sample 15 positive cases and 10 negative cases to form an abductive instance. The negative cases are constructed by randomly changing the orders in positive cases. 

\end{itemize}
\end{itemize}

\begin{figure*}[t]
\centering
\includegraphics[width=0.65\linewidth]{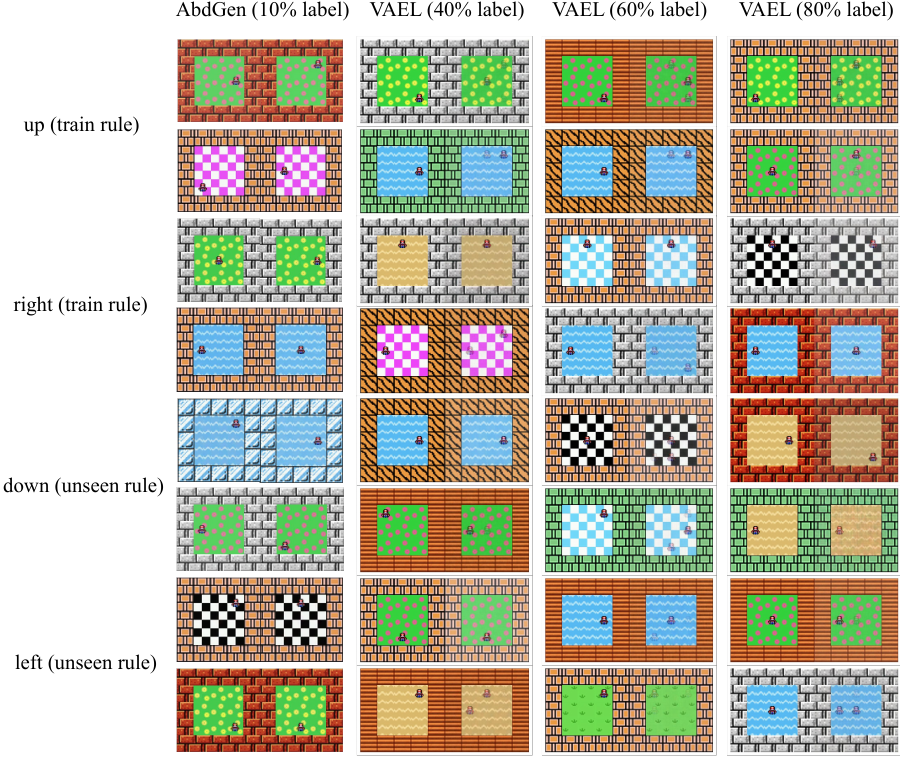}
\caption{Qualitative results for the generation under unseen rules experiment in comparing AbdGen with VAEL. "up (train rule), right (train rule)" indicate the generation results on training-stage rules. "down (unseen rule), left (unseen rule)" indicate the results on unseen rules during training.} 
\label{fig:exp_newrulegen}
\end{figure*}
\begin{figure*}[t]
    \centering
        \includegraphics[width=1\linewidth]{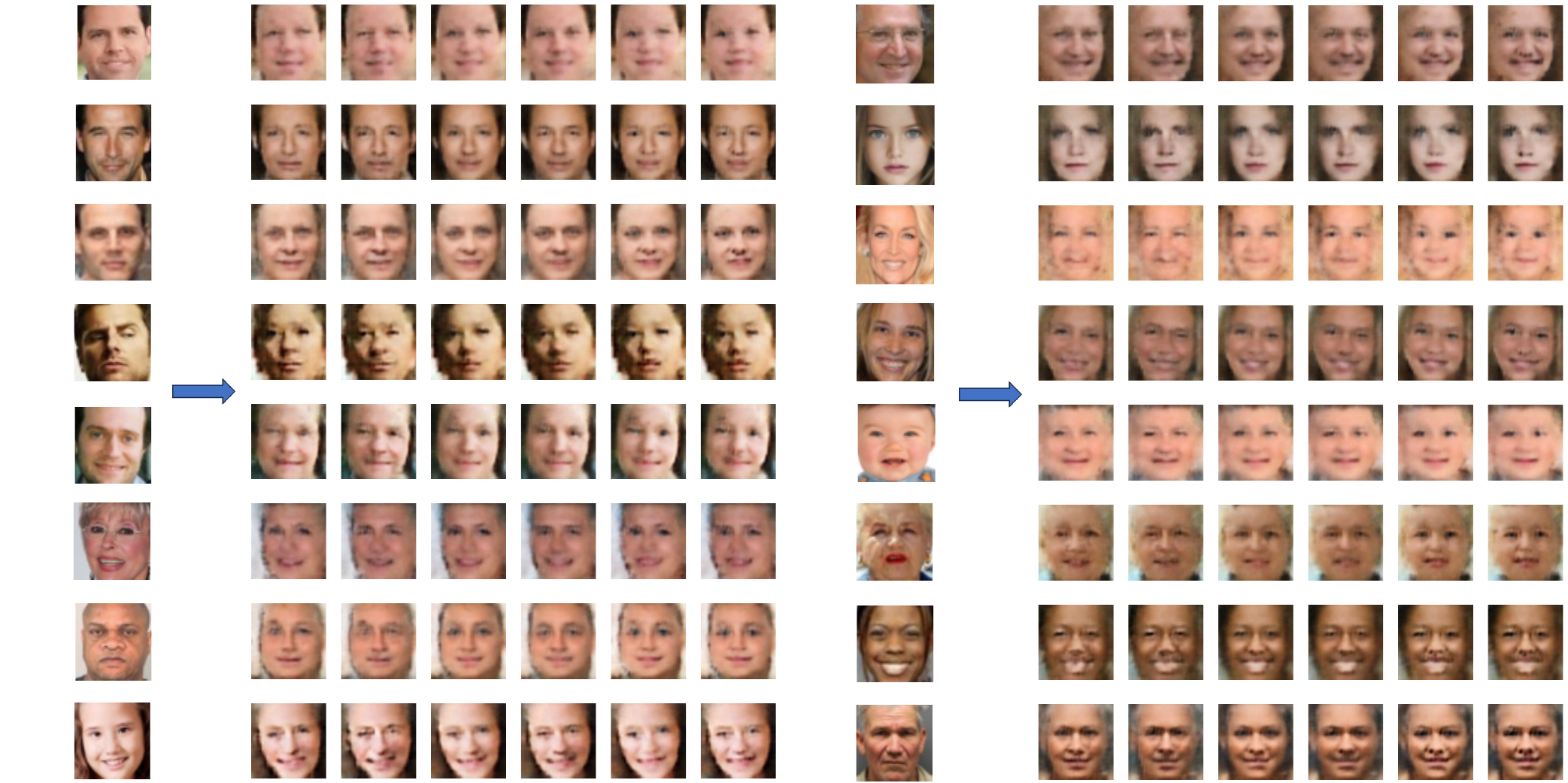}
    \caption{More qualitative generation results for the learning with rule induction experiment on UTKFace.}
    \label{fig:moreface}
\end{figure*}

\end{document}